\documentclass{article}

\usepackage{arxiv}

\usepackage[utf8]{inputenc} 
\usepackage[T1]{fontenc}    
\usepackage{hyperref}       
\usepackage{url}            
\usepackage{booktabs}       
\usepackage{amsfonts}       
\usepackage{nicefrac}       
\usepackage{microtype}      
\usepackage{amsmath}
\usepackage{cleveref}       
\usepackage{lipsum}         
\usepackage{graphicx}
\usepackage{natbib}
\usepackage{doi}
\usepackage{graphicx}

\usepackage{amssymb}
\usepackage{booktabs}
\usepackage{algorithm}
\usepackage{algorithmic}
\usepackage{bbm}

\usepackage{amsmath,amsfonts,bm}

\def\eqref#1{equation~\ref{#1}}

\def\1{\bm{1}}

\def\ve{{\bm{e}}}

\def\vt{{\bm{t}}}

\DeclareMathAlphabet{\mathsfit}{\encodingdefault}{\sfdefault}{m}{sl}
\SetMathAlphabet{\mathsfit}{bold}{\encodingdefault}{\sfdefault}{bx}{n}

\def\sP{{\mathbb{P}}}

\newcommand{\R}{\mathbb{R}}

\title{Class Enhancement Losses with Pseudo Labels for \\Zero-shot Semantic Segmentation}

\author{ Son Duy Dao\\
	Faculty of Information Technology\\
	Monash University\\
	\texttt{duy.dao@monash.edu} \\
	\And
	Hengcan Shi\\
	Faculty of Information Technology\\
	Monash University\\
	\texttt{hengcan.shi@monash.edu} \\
        \And
        Dinh Phung\\
	Faculty of Information Technology\\
	Monash University\\
	\texttt{dinh.phung@monash.edu} \\
        \And
        Jianfei Cai\\
	Faculty of Information Technology\\
	Monash University\\
	\texttt{jianfei.cai@monash.edu} \\
}

\begin{document}
\maketitle

\begin{abstract}
Recent mask proposal models have significantly improved the performance of zero-shot semantic segmentation.  
However, the use of a `background' embedding during training in these methods
is problematic as the resulting model tends to over-learn and assign all unseen classes as the background class instead of their correct labels. 
Furthermore, they ignore the semantic relationship of text embeddings, which arguably can be highly informative for zero-shot prediction as seen classes may have close relationship with unseen classes. 
To this end, this paper proposes novel class enhancement losses to bypass the use of the background embbedding during training, and simultaneously exploit the semantic relationship between text embeddings and mask proposals by ranking the similarity scores. To further capture the relationship between seen and unseen classes, we propose an effective pseudo label generation pipeline using pretrained vision-language model.
Extensive experiments on several benchmark datasets show that our method achieves overall the best performance for zero-shot semantic segmentation. 
Our method is flexible, and can also be applied to the challenging open-vocabulary semantic segmentation problem.
\end{abstract}


\section{Introduction}

Semantic segmentation, grouping image pixels into the semantic categories, is a fundamental task in computer vision. Although showing a huge progress recently \cite{cheng2021per, cheng2022masked, yu2022cmt, zhou2022rethinking, li2022deep, wang2021max, xie2021segformer, zhu2021unified, zhu2021unified}, the performance is still limited to a fixed set of categories and the quality of human annotations. In order to mitigate the great burden of human segmentation annotations, zero-shot algorithms \cite{bucher2019zero, xu2021simple, ding2022decoupling, xu2022groupvit, kato2019zero, zhang2021prototypical} have been proposed, aiming to train segmentation models with less annotations. Specifically, the goal of zero-shot semantic segmentation (ZS3) is to segment novel categories with no corresponding mask annotations during training.

\begin{figure}[t!]
    \centering
    \includegraphics[width=\columnwidth]{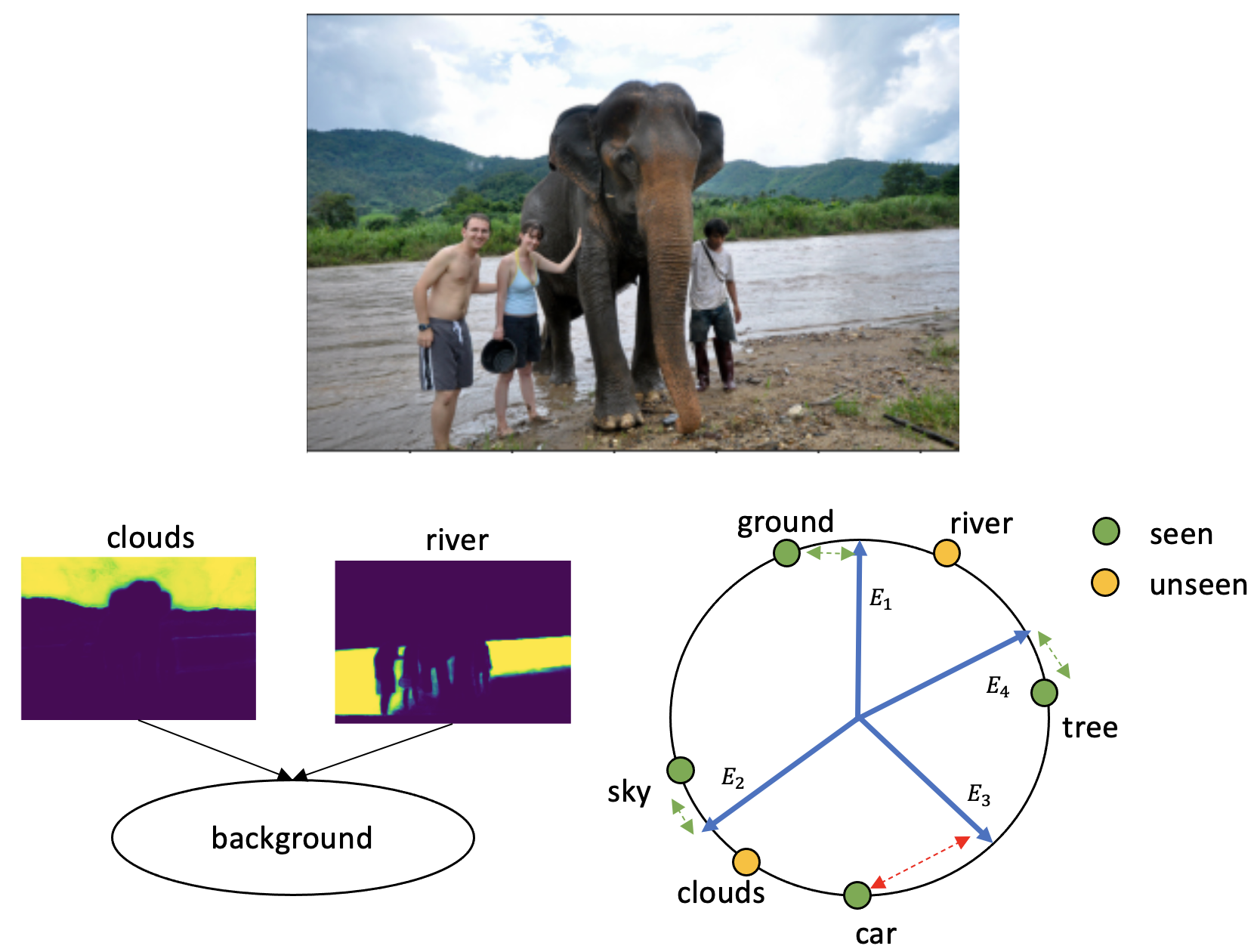}
    \caption{Current mask-proposal-based zero-shot segmentation models have two limitations. First, multiple unseen categories (e.g., `clouds', `river') are classified as the same `background' class. As a consequence, the learned model is biased toward seen classes. Second, label relationship is not exploited. Seen labels and unseen labels may co-exist (e.g. `river' or `tree' often co-exist with `sky, ground'). The model trained on the seen class `sky' may generalize well to the unseen class `clouds' due to their close semantic relation. 
    }
    \label{fig:intro}
\end{figure}

Early ZS3 works \cite{bucher2019zero, hu2020uncertainty} rely on fully-convolutional-network-based (FCN-based) segmentation models \cite{long2015fully}, which typically model semantic segmentation as a pixel-wise classification problem and treat text embeddings as classifiers to classify every pixel. The latest ZS3 works 
\cite{ding2022decoupling, xu2021simple} are based on mask proposals \cite{cheng2021per}, which divides the ZS3 problem into two sub-tasks: first predicting a set of class-agnostic mask proposals, and then classifying each mask proposal into a specific category. As demonstrated in \cite{xu2021simple}, the model trained on only mask annotations of seen classes produces high pixel accuracy for class-agnostic mask predictions, which means it can well segment out the objects/ concepts for unseen classes. Further, powerful vision-language pretrained models can be naturally combined to classify mask proposals. The above two points lead to the great performance improvement for ZS3. 

Despite the improvement over the traditional pixel-wise classification methods, one major limitation of the current mask-proposed-based ZS3 works is that they
still use a trainable `background' embedding similar to the fully supervised setting. Specifically, as shown in Figure~\ref{fig:intro} (left), 
during training, although the model can segment different objects of unseen classes, it treats all these novel objects/concepts into the same `background' class with the cross-entropy loss. 
%
This is inappropriate because unseen classes can be very different in semantic meaning (e.g `river' and `clouds'). As a consequence, during inference, the model tends to have two behaviors: either classify all mask proposals of unseen classes into the `background' class instead of their correct labels, or bias to seen classes instead of predicting for unseen classes because the classification scores for unseen classes are always low. Therefore, it requires a proper treatment for `background' proposals for the mask-proposal-based ZS3 solutions. The `background' proposals should not affect the predictions of seen classes, and they need to be semantically meaningful as they may belong to different unseen classes. 

Motivated from the above observations, in this paper, we introduce novel training losses to tackle ZS3. Our method is also based on the mask proposal approach, but we make two improvements in the training losses. Firstly, to mitigate the problem caused by `background' embedding, we avoid using the `background' embedding during training. Instead, by viewing the mask prediction for a class as the weighted sum of all mask proposals, we propose to constrain `background' proposals so that they contribute equally for all classes.
In this way, `background' proposals will not affect the prediction of other classes. Secondly, considering seen classes might have strong semantic relationships with some unseen classes (e.g., `sky' and `clouds'), 
we propose to use a multi-label ranking loss to incorporate the semantic relationship into mask proposals. The multi-label ranking loss compares the similarity scores between mask proposals and the set of label embeddings, and then a set of relevance labels are ranked higher than a set of irrelevance labels. As a result, it helps the model extract semantic meaningful mask proposals which have high similarity scores with the relevance labels. This relevance label set might contain the unseen classes. To further enhance the co-orccurence relationship between seen and unseen labels, we propose to use pretrained vision-language (e.g. CLIP \cite{radford2021learning}) to generate pseudo unseen labels for each image. Since the label relationship is exploited, the trained model on a predefined set of labels can be effectively evaluated on another set of labels without any adaptation.

Our contributions can be summarized as follows:
\begin{itemize}
    \item By viewing the mask predictions as a weighted sum of all mask proposals, we propose to remove the `background' embedding and constrain background proposals to contribute equally to all classes. In this way, background mask proposals will not affect the mask predictions of ground-truth classes.
    \item We incorporate semantic relationship into mask proposals by using a multi-label ranking loss. To further enhance the relationship between seen and unseen labels, we propose a pseudo label generation pipeline for each image.
    \item Experiments on several benchmark datasets show that our approach significantly improves the performance of GZS3. It also transfers well to different datasets on open-vaocabulary semantic segmentation setting.
\end{itemize}

\section{Related Works}

\subsection{Zero shot semantic segmentation.}
ZS3 aims to segment novel classes that are not annotated during training. Traditionally, ZS3 is often modeled as pixel-wise classification. For example, SPNet \cite{xian2019semantic} learns a segmentation network that projects intermediate pixel-feature maps into the word embedding space, and then produces class probabilities via a fixed word embedding projection matrix. 
In the same spirit, \cite{baek2021exploiting} proposes to learn a joint embedding space for the visual encoder and the semantic encoder.
 Other approaches use generative models to synthesize visual features of unseen classes from corresponding word embeddings, and then train the segmentation model with both real and synthesized features. A typical work for this line is \cite{bucher2019zero}. Later, CSRL \cite{li2020consistent} further incorporates the semantic relation between seen and unseen features. CaGNet \cite{gu2020context} utilizes contextual information and SIGN \cite{cheng2021sign} exploits spatial information for feature generation. Some other works focus on different aspects of ZS3. In \cite{hu2020uncertainty}, the uncertainty losses are proposed to reduce the effect of outline samples.
 \cite{pastore2021closer} carefully studies self-training technique for zero-shot semantic segmentation. 
 \cite{lv2020learning} explores the transductive learning setting for ZS3.
 However, these pixel-wise classification based methods have two limitations: (1) Word embeddings are for describing objects, which are not robust to classify pixel features; (2) Since they are early works, vision-language pre-training models have not been used.
 
 Recent methods utilize the large-scale vision-language pre-trained model CLIP \cite{radford2021learning} to boost the performance of ZS3. For instance, \cite{zhou2021denseclip} extracts meaningful localization information for unseen classes from CLIP as pseudo masks to train the model in a self training fashion. \cite{li2022language} learns an image encoder by pixel-wise classification using text embeddings from CLIP text encoder. In another direction, a two-stage mask proposal framework has been proposed in \cite{xu2021simple}. In the first stage, a mask proposal model is trained to extract class-agnostic mask for objects in the images, and then these segmentation proposals are classified in the second stage. \cite{ding2022decoupling} further extends this framework to an end-to-end fashion. While showing great performance, these works often use a trainable `background' embedding to represent unseen classes, which makes the model bias to seen classes during testing as all unseen classes are classified as `background'. 
 In our work, we exploit the relationship of weak labels from both seen and unseen sets. We also avoid the usage of the `background' embedding by classifying background proposals equally to all the classes.
 
 \subsection{Open-Vocabulary Semantic Segmentation}
 Open-Vocabulary Semantic Segmentation (OVS) is an emerging problem where the model is trained and evaluated on different datasets with diverse and arbitrary label sets. Previous GZS3 methods \cite{xu2021simple, ding2022decoupling} can be applied to this setting, but the performances are not high because of the aforementioned limitations. Recent method \cite{ghiasi2021open} proposes to align visual segment and text embedding using visual grounding loss. Unlike these works, we exploit the label co-occurrence relationship to improve performance.
 
 \subsection{Multi-label Ranking.}
 The multi-label ranking loss is a common loss function used in multi-label zero-shot learning \cite{zhang2016fast, huynh2020shared, narayan2021discriminative}. Its goal is to train a model to extract a set of principal vectors from an input image so that when using those vectors to rank a set of label embeddings, the set of relevance labels to the image should have a higher rank than the set of irrelevance labels. 
 In our works, we view mask proposals as principal vectors extracted from the input image and leverage the multi-label ranking loss to learn the label relationship for the masks.
 
 \begin{figure*}[ht!]
    \centering
  \includegraphics[width=\textwidth]{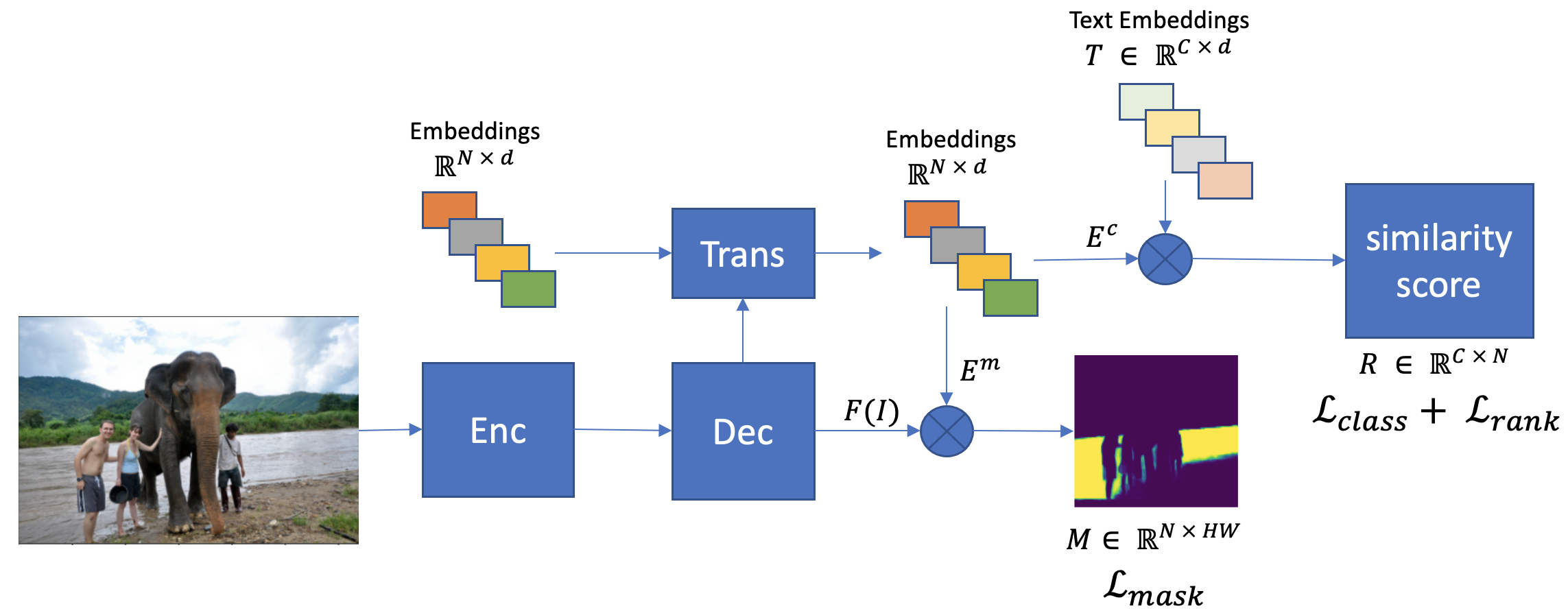}
  \caption{Overview of our proposed method, which follows the mask proposal family. It learns a set of $N$ embeddings via a Transformer decoder (`Trans'), which is projected into two forms, $E^c$ and $E^m$, for the class branch (top) and the mask branch (bottom), respectively. In the mask branch,  $E^m$ is multiplied with pixel feature map $F(I)$ from the image decoder to obtain mask proposals $M$, which are trained with loss $\mathcal{L}_{mask}$. In the class branch, class embeddings $E^c$ are multiplied with text embeddings $T$ to get similarity score matrix $R$, which are trained with our proposed background-aware class loss $\mathcal{L}_{class}$ and ranking loss $\mathcal{L}_{rank}$. }
  \label{fig:overview}
\end{figure*}

\section{Method}
In this section, we first introduce the problem definition,  then describe our mask-proposal-based method for zero-shot semantic segmentation, and finally present our proposed loss functions.

\subsection{Problem definition}
The goal of zero-shot semantic segmentation is to predict masks for unseen categories $U$ by learning from mask annotations of seen categories $S$, where the seen categories and unseen categories are disjoint, i.e., $S \cap U = \varnothing$. Typically the label set consists of semantic words (e.g., cat, dog, car, etc) or their definitions. Therefore, the model learned on seen categories can implicitly exploit semantic relationships of word embeddings to predict on unseen categories.
%
There are two common settings for zero-shot semantic segmentation: 
\begin{itemize}
    \item Zero-shot semantic segmentation (ZS3): train on mask annotations of $S$, and test on $U$.
    \item Generalized zero-shot semantic segmentation (GZS3): train on mask annotations of $S$ and test on $S \cup U$, 
\end{itemize}
where GZS3 is a more common one. Depending on whether the unlabeled pixels are observed during training, the setting can be split into inductive setting (not observed) or transductive setting. We follow the transductive GZS3 works \cite{pastore2021closer, zhou2021denseclip} to use image-level labels, when unseen class labels are pseudo labels. 
Note that previous works \cite{xu2021simple, ding2022decoupling, pastore2021closer, zhou2021denseclip}  have to train the model on paired masks and labels (i.e., which mask belongs to which label), since the pair information is important for matching predictions and ground truths during training. In our work, we do not need this paired information by using unpaired class-agnostic masks and labels. This allows us to use both seen and unseen categories (class name) as training labels, while only mask annotations of seen categories are needed. The benefit can be further extend to OVS setting, where the model is trained once and can be evaluated on different datasets  without any dataset-specific adaptation.

\subsection{Mask-proposal-based model for GZS3}
\textbf{Overview.} We follow the recent popular pipeline \cite{xu2021simple, ding2022decoupling}, which decouple GZS3 problem into two sub-problems: learning a model to do class-agnostic segmentation, and concurrently classify the class-agnostic masks into suitable categories. 


Particularly, our method is built on a mask-proposal-based model (e.g.,   Mask2Former~\cite{cheng2021per}). Fig.~\ref{fig:overview} shows the overall architecture, where $N$ segment-level $d$-dimensional embeddings of are learned from the image feature map via a Transformer decoder. These $N$ embeddings are then passed through a class projection layer and a mask projection layer to generate class embddings $E^c \in \R^{N \times d}$ and mask embeddings $E^m \in \R^{N \times d}$, respectively. 

\noindent\textbf{Mask prediction.} Let the image feature map be denoted as $F(I) \in \R^{d \times H \times W}$, where $H$ and $W$ are the height and width of the feature map. The mask prediction (Fig.~\ref{fig:overview} bottom) for each mask embedding is computed as $M_q = \sigma(\ve^m_q \cdot F(I)) \in \R^{H \times W}$, where $\sigma$ is Sigmoid function and $\ve^m_q \in E^m, q \in [1,...,N]$, is the $q$-th mask query in $E^m$. Then, we obtain all mask proposals $M =\{M_q| q \in [1,...,N]\}$. For simplicity, we denote $M$ as a matrix of size $N \times HW$. Note that $N$ is usually smaller than the number of classes $C$, where $C = |S \cup U|$.

The mask predictions are trained with a binary mask loss. Specifically, Hungarian-matching algorithm \cite{kuhn1955hungarian} is first applied to obtain a one-by-one matching between mask predictions and class-agnostic mask ground truths. 
Then, the mask loss is calculated as a combination of focal loss \cite{lin2017focal} and DICE coefficient loss \cite{milletari2016v}.
This mask loss is widely used in the mask-proposal-based semantic segmentation family \cite{cheng2021per, cheng2022masked}, and it 
helps the model produce meaningful segmentation regions without caring much about semantic information. Therefore, it can be used to segment objects of unseen classes. 

\noindent\textbf{Class Prediction.} For the class prediction shown in Fig.~\ref{fig:overview} top, we extract text embeddings for each `class name' in $S \cup U$ via a pretrained text encoder (e.g., CLIP \cite{radford2021learning}), denoted as $T = \{\vt_i \in \R^d, i \in [1,...,C]\}$, where $\vt_i$ is the text embedding for the $i$-th class. 
%
We then compute a similarity score matrix $R \in \R^{C \times N}$ between $C$ text embeddings $T$ and $N$ class embeddings $E^c$. Each element in the similarity score matrix is a cosine similarity between one class embedding $\ve^c_q \in E^c$ and one text embedding $\vt_i \in T$: $r_{qi} = \frac{\ve^c_q \cdot \vt_i}{|\ve^c_q| \cdot |\vt_i|}$. Hungarian-matching is also applied to 
get a one-to-one matching between class embeddings and ground-truth class labels with the similarity score being the cost. 
The cross-entropy loss for classification is computed as follows:
\begin{align}
    \mathcal{L}_{class} &= \mathcal{L}_{CE} = -\sum_{q=1}^N \text{log} p_{q}(c_q^{gt})\\
    p_q(c_i) &= \frac{exp(r_{qi}/\tau)}{\sum_{j=0}^C exp(r_{qj}/\tau)}
\end{align}
where $c_q^{gt}$ is the matched ground truth label for class embedding $\ve^c_q$, and $\tau$ is the temperature. Note that in the previous works \cite{xu2021simple, ding2022decoupling}, a randomly initialized `background' text embedding $\vt_0 \in \R^d$ is often used to represent the background class, which is not associated to any of the training class labels. This `background' text embedding is learned during training the whole network and the class embeddings that do not have corresponding ground truth masks are considered `background' class, which causes the aforementioned bias problem. We will address this issue by our background-aware class loss later.
%

During training, only the pixels belonging to seen classes $S$ are used for the mask loss, and both $S$ and $U$ are used to train the classification loss. We ignore any mask predictions that do not have matched mask ground-truths. The loss for the entire model is a weight sum of the class-agnostic mask loss $\mathcal{L}_{mask}$ and the classification loss $\mathcal{L}_{class}$:
\begin{align}
    \mathcal{L}_{total} = \alpha \mathcal{L}_{class} + \beta \mathcal{L}_{mask}
\end{align}
where $\alpha$ and $\beta$ are trade-off weights.

\begin{figure}[t]
    \centering
    \includegraphics[width=\columnwidth]{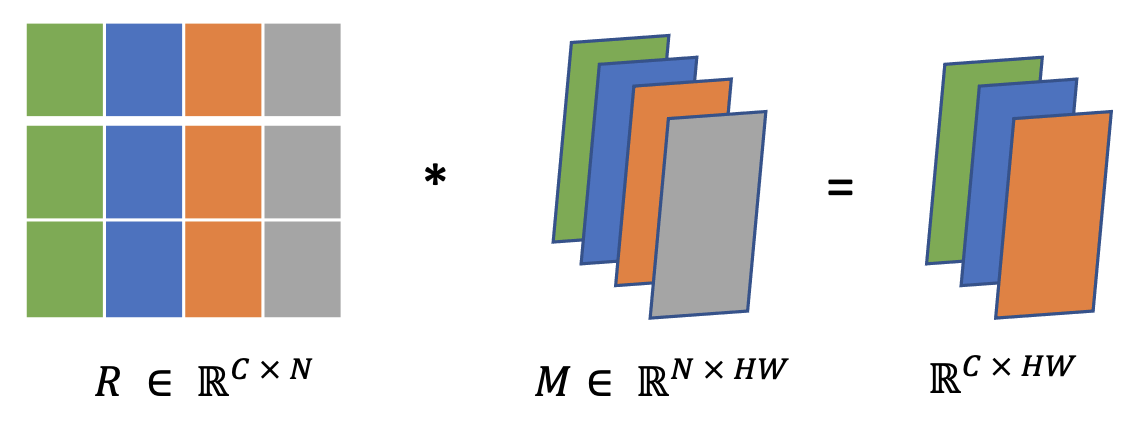}
    \caption{The inference of semantic segmentation during testing is based on the matrix multiplication of similarity score matrix $R$ and mask proposals $M$. The mask prediction for a class $c$ is the weighted sum of all mask proposals, where the weights are the similarity scores of class $c$ with $N$ class embeddings. 
    }
    \label{fig:mask_infer}
\end{figure}

\noindent\textbf{Inference.} Fig.~\ref{fig:mask_infer} illustrates the inference process. In particular, after obtaining the classification score and the mask prediction for each query, we compute the final mask for each class as 
\begin{align}
    \underset{c = S + U}{\arg\max} \sum_{q=1}^N p_q(c)M_q[h,w].
\end{align}
It indicates that the output mask for a class $c$ is the weight sum of $N$ proposal masks where the weights are the similarity scores of class $c$ over $N$ class embeddings. Thus, a mask proposal contributes the most for a class if its corresponding similarity score for that class is the highest. 

\subsection{Class Enhancement Losses}
\noindent\textbf{Background-aware class loss.} While using the `background' classifier is a common practise in the fully supervised semantic segmentation setting, it may harm in ZS3 settings. This is because foreground objects that do not associated with any ground truth masks often belong to one of unseen classes, and using one single `background' class is hard to represent various unseen classes. Therefore, in this research, we propose to remove the background text embedding and consider the class embeddings that do not match to any ground truth class labels as `background'.

Considering the output mask for a class is the sum of all mask proposals weighted by the similarity scores of that class over N class embeddings. To avoid the `background' class embeddings affect other non-background class masks, our idea is to encourage their classification prediction to be uniform over all classes. Specifically, we define our background-aware classification loss as
\begin{align}
    \mathcal{L}^{bg}_{class} = \lambda \mathbbm{1}_{q \in H} \mathcal{L}_{CE} + (1-\lambda)\mathbbm{1}_{q \notin H} KL(p_q||D_U)
\end{align}
where $H$ is the set of class embeddings with matched ground truth labels, $\mathbbm{1}$ is the indicator function, $KL$ is  the Kullbank-Leibler (KL) divergence loss, $p_q$ is the classification prediction, $D_U$ represents uniform distribution and $\lambda$ is a hyper-parameter. 

%

\noindent\textbf{Multi-Label Ranking Loss.} Previous methods often treat each class query separately, neglecting the class relationship which can be beneficial for zero-shot prediction. For example `sheep' and `grass' is more likely to co-occur than 'sheep' and 'airplane' in semantic space. 
Thus, we propose to further use multi-label ranking loss to implicitly introduce that semantic structure into the classification. Specifically, following \cite{zhang2016fast}, the ranking loss function for one image $i$ is defined as
\begin{align} \label{eq:lrank}
    \mathcal{L}^{i}_{rank} = \frac{1}{|\sP^i|}\sum\limits_{j \in \sP^i}\sum\limits_{k \in
    \Bar{\sP^i}}\log(1+\exp{(r^*_{\cdot k} - r^*_{\cdot j})})
\end{align}
where $\sP^i$ and $\Bar{\sP}^i$ are the sets of positive labels and negative labels of image $i$, respectively, and $r^*_{\cdot k}$ is the largest similarity score of class $k$ over all $N$ class embeddings, which can be obtained from the similarity score matrix $R \in \R^{C \times N}$. Minimizing Eq.~(\ref{eq:lrank}) essentially encourages the max similarity score of any positive label to be higher than the max similarity score of any negative label.
%
%
The total ranking loss is the mean over all images within a mini-batch $B$:
\begin{align}
    \mathcal{L}_{rank} = \frac{1}{B}\sum\limits_{i \in B}\mathcal{L}^{i}_{rank}
\end{align}
Finally, our total training loss becomes
\begin{align}
    \mathcal{L}_{total} = \alpha \mathcal{L}^{bg}_{class} + \beta \mathcal{L}_{mask} + \gamma \mathcal{L}_{rank}.
\end{align}


\subsection{Generate Pseudo Unseen Labels.}
Since the ranking loss captures the co-occurrence relationship between labels, training the model with only seen labels cannot fully capture the relationship between seen and unseen labels. Therefore, we design a simple way to assign pseudo unseen labels for each image. The goal is for each image, we obtain a predicted score vector $\hat{s} \in R^U$ for all $U$ unseen classes. Then we threshold this vector score to obtain the pseudo unseen labels.

For each image with unannotated pixels, we use a pretrained class-agnostic Mask2Former \cite{cheng2022masked} model to extract $N$ mask proposals. For each mask proposal, we infer the bounding box that covers the whole mask and crop the input image to get $N$ cropped images. The cropped images are fed through the CLIP\cite{radford2021learning} image encoder to get $N$ image embeddings $\hat{E} \in R^{N \times d}$, where $d$ is the embedding's dimension. Assume that $\hat{T} = \{\hat{\vt}_j \in \R^d, j \in [1,...,U]\}$ are the text embeddings of unseen classes extracted from CLIP text encoder. We compute the classification score $\hat{S} = \{\hat{s}_i \in R^U, i \in [1,...,N]\}$ of the image embeddings $\hat{E}$, where the classification score for the $i^{th}$ image embedding $\hat{s}_i$ is computed as follows:
\begin{align}
    \hat{s}_i = softmax(cos(\hat{e}_i, \hat{T})), i \in N
\end{align}
where $cos(\hat{e}_i, \hat{T})$ is the cosine similarity between $\hat{e}_i$ and each text embedding in $\hat{T}$. The final score for each class $j \in U$ is defined as the maximum score of class $j$ over $N$ image embeddings:
\begin{align}
    \hat{s}_j = \underset{i = [1, .., N]}{\max}(\hat{s}_{ij}); j \in U
\end{align}
The predicted classification scores of all unseen classes for the input image are $\hat{s} = \{\hat{s}_1,..., \hat{s}_U\}$. Finally, we threshold these classification scores with a value $\tau = 0.99$ to obtain the pseudo unseen labels.

\section{Experiments}
\subsection{Datasets}
We use several well-known datasets to train and evaluate our proposed method, including COCO-Stuff, ADE20K, Pascal Context, and Pascal VOC.

\noindent\textbf{COCO-Stuff} \cite{caesar2018coco} is a large scale dataset containing mask annotations for 171 classes including things (e.g. dog, cat) and stuffs (e.g. grass, sky). The dataset contains over 118K training images and 5K validation images. According to \cite{xian2019semantic}, the annotations are further divided into 156 seen classes and 15 unseen classes for zero-shot setting. 

\noindent\textbf{Pascal VOC 2012 (VOC-20)} \cite{everingham2010pascal} contains 20 classes in total and is split into 15 seen classes and 5 unseen classes. It has 11185 training images and 1449 validation images. Following other works \cite{bucher2019zero, cheng2021sign}, we also use the augmented annotations in \cite{BharathICCV2011} for training.

\noindent\textbf{Pascal Context} \cite{mottaghi2014role} is an extension from VOC-20 where the annotations are densely created for each image. 
There are 5105 images for testing with two versions: one version uses the most 59 frequent classes (\textbf{P-59}), and the other version uses full 459 classes (\textbf{P-459}).

\noindent\textbf{ADE20K} \cite{zhou2017scene} contains 2000 validation images. We use two versions: one with most frequent 150 classes (\textbf{A-150}) and the other with diverse 847 classes (\textbf{A-847}).

\subsection{Evaluation Metrics}
Following previous works \cite{xu2021simple, ding2022decoupling}, we adopt three metrics. The first is mIoU(S), which is the mean of class-wise intersection over union (mIoU) for seen classes. The second is mIoU(U), meaning the mIoU for unseen classes. The third is the harmonic mean IoU (hIoU) between both seen and unseen classes, which is computed as:
\begin{align}
    hIoU = \frac{2 \times mIoU_{seen} \times mIoU_{unseen}}{mIoU_{seen} + mIoU_{unseen}}
\end{align}
We follow others to evaluate our model on GZS3 setting. 

\subsection{Implementation Details}
We conduct all experiments on two Nvidia 2080ti GPUs. We train a Mask2Former \cite{cheng2022masked} model on the COCO Stuff dataset with ResNet-50 \cite{he2016deep} as the default backbone. To accelerate training, for each dataset, the backbone is first trained in fully-supervised setting for seen classes. Then, we use this pretrained backbone to train our proposed method in GZS3 setting. The weight for each loss component is empirically set to $\alpha=2, \beta=5, \gamma=1$ and $\lambda=0.6$. 
100 queries for both training and testing are used.
We also use an AdamW optimizer \cite{loshchilov2017decoupled} with the initial learning rate of 5e-5, weight decay of 1e-4 and a backbone multiplier of 0.1.  On the COCO Stuff dataset, the batch size is set to 8 and the total training iteration is 50K with two learning rate decay steps at 5K and 80K.  On the VOC-20 dataset, we use batch size 8 with a total training iterations of 5K, and keep all other setting as the same as COCO Stuff.
For all other settings and hyper-parameters, we keep the official setting of \cite{cheng2022masked} without changes. 

For the pretrained CLIP text encoder, the ViT-B/16 \cite{dosovitskiy2020image} backbone is used as default if not specified. We also use prompt templates with class names for all experiments. 
Prompt templates are listed in the supplementary file.

\subsection{Zero-shot Semantic Segmentation.}
\begin{figure*}[ht!]
    \centering
  \includegraphics[width=\textwidth]{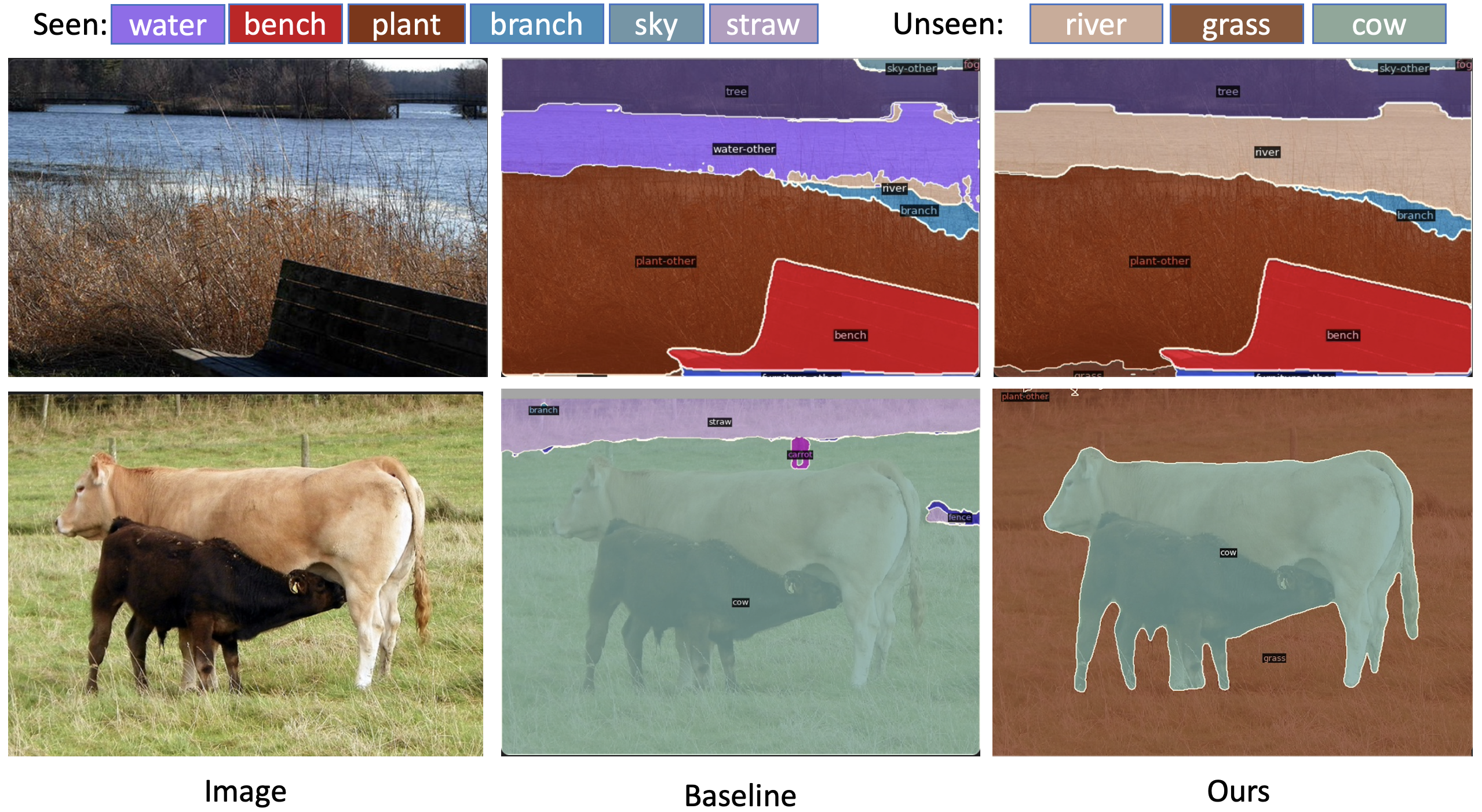}
  \caption{Visualization comparisons between the baseline and our proposed method. The baseline fails to classify some unseen classes (i.e., `river' or `grass'), while our method successfully classifies these regions. Images from COCO Stuff dataset.}
  \label{fig:vis1}
\end{figure*}
We compare our method with previous state-of-the-art pixel-wise methods and mask-proposal methods (e.g., SPNet \cite{xian2019semantic}, ZS3 \cite{bucher2019zero}, CaGNet \cite{gu2020context}, SIGN \cite{cheng2021sign}, Joint \cite{baek2021exploiting}, Zegformer \cite{ding2022decoupling}, Simple \cite{xu2021simple} and STRICT \cite{pastore2021closer}) on COCO-Stuff and VOC-20 datasets. We report the results of compared methods from their papers. Some works use self-training techniques, where we mark them as `ST'. 

Results on the COCO Stuff dataset are shown in Table~\ref{tab:coco}. Our method achieves 5.7\% improvements for unseen classes, compared with the second best method (Simple). Our model also obtains a competitive mIoU for seen classes and the best hIoU for both seen and unseen classes.

\begin{table}[t!]
    \centering
    \begin{tabular}{|c|c|c|c|c|}
    \hline
       \textbf{Method}  & \textbf{ST} & \textbf{mIoU(S)} & \textbf{mIoU(U)} & \textbf{hIoU} \\
       \hline
       \multicolumn{5}{|l|}{\emph{Pixel-wise methods}}\\
       \hline
        SPNet \cite{xian2019semantic} & x & 35.2 & 8.7 & 14.0\\
        ZS3 \cite{bucher2019zero} & x & 34.7 & 9.5 & 15.0\\
        CaGNet \cite{gu2020context} & x & 33.5 & 12.2 & 18.2\\
        SIGN \cite{cheng2021sign}& x & 32.3 & 15.5 & 20.9\\
        \hline
        ZS3 \cite{bucher2019zero} & \checkmark & 34.9 & 10.6 & 16.2\\
        CaGNet \cite{gu2020context} & \checkmark & 35.6 & 13.4 & 19.5\\
        SIGN \cite{cheng2021sign}& \checkmark & 31.9 & 17.5 & 22.6\\
        SPNet \cite{xian2019semantic} & \checkmark & 34.6 & 26.9 & 30.3\\
        STRICT \cite{pastore2021closer} & \checkmark & 35.3 & 30.3 & 32.6\\
        \hline
        \multicolumn{5}{|l|}{\emph{Mask-proposal methods}}\\
        \hline
        Zegformer \cite{ding2022decoupling} & x & 36.6 & 33.2 & 34.8\\
        Simple \cite{xu2021simple}  & x & \textbf{39.3} & 36.3 & 37.8\\
        \textbf{Ours} & x & 38.6 & \textbf{42.0} & \textbf{40.2}\\
    \hline
    \end{tabular}
    \caption{GZS3 results on the COCO Stuff dataset. `ST' stands for the self-training technique.}
    \label{tab:coco}
\end{table}

Table~\ref{tab:voc12} shows the results on the VOC-20 dataset. The proposed method outperforms previous works by 5.0\%, 2.3\% and 3.6\% for seen classes, unseen classes and the comprehensive hIoU metric, respectively. Without self-training, we also achieve the best performance on all metrics when compare to pixel-wise methods.


\begin{table}[t]
    \centering
    \begin{tabular}{|c|c|c|c|c|}
    \hline
       \textbf{Method}  & \textbf{ST} & \textbf{mIoU(S)} & \textbf{mIoU(U)} & \textbf{hIoU} \\
       \hline
       \multicolumn{5}{|l|}{\emph{Pixel-wise methods}}\\
       \hline
        SPNet \cite{xian2019semantic} & x & 78.0 & 15.6 & 26.1\\
        ZS3 \cite{bucher2019zero}& x & 77.3 & 17.7 & 28.7\\
        CaGNet \cite{gu2020context}& x & 78.4 & 26.6 & 39.7\\
        SIGN \cite{cheng2021sign}& x & 75.4 & 28.9 & 41.7\\
        Joint \cite{baek2021exploiting} & x & 77.7 & 32.5 & 45.9\\
        \hline
        ZS3 \cite{bucher2019zero}& \checkmark & 78.0 & 21.2 & 33.3\\
        CaGNet \cite{gu2020context} & \checkmark & 78.6 & 30.3 & 43.7\\
        SIGN \cite{cheng2021sign}& \checkmark & 83.5 & 41.3 & 55.3\\
        SPNet \cite{xian2019semantic}& \checkmark & 77.8 & 25.8 & 38.8\\
        STRICT  \cite{pastore2021closer}& \checkmark & 82.7 & 35.6 & 49.8\\
        \hline
        \multicolumn{5}{|l|}{\emph{Mask-proposal methods}}\\
        \hline
        Zegformer \cite{ding2022decoupling} & x & 86.4 & 63.6 & 73.3\\
        Simple \cite{xu2021simple}  & x & 83.5 & 72.5 & 77.5\\
        \textbf{Ours} & x & \textbf{88.5} & \textbf{74.8} & \textbf{81.1}\\
        \hline
    \end{tabular}
    \caption{GZS3 results on Pascal VOC (VOC-20). `ST' stands for the self-training technique.}
    \label{tab:voc12}
\end{table}



\subsection{Open-vocabulary semantic segmentation.}

\noindent\textbf{Experimental Setup.} Our method can be easily extend into the challenging OVS setup. The model only need to be trained once and can be applied to segment various datasets with diverse number of classes. Specifically, we train our model using COCO-Stuff mask annotation of 156 seen classes, and nouns extracted from COCO-Caption \cite{chen2015microsoft}. To reduce the noise from caption annotation, we only choose in total 1047 most frequent nouns for training (including 156 seen class names). We randomly drop each words by a probability of 0.25 to prevent the model from overfitting to the noise ground truth. The trained model is directly evaluated on new datasets including A-847, A-150, P-459, P-59 and VOC-20 without any dataset-specific adaptation.

\noindent\textbf{Main Results.} Table \ref{tab:open_vocab} shows the mIoU performance of our method on OVS compare with several other works, including per-pixel segmentation approach (e.g LSeg+) and mask proposal approaches (e.g. Simple, Zegformer, and OpenSeg). Among the mask proposal approaches, we can see that our approach achieves the best performance on all five evaluated datasets. Specifically, our model trained on COCO-Caption's nouns outperforms OpenSeg trained on extra labels from Localized Narrative \cite{pont2020connecting} by 0.2\%, 1.7\%, 1.9\%, 7.9\%, 22.1\% on A-847, P-459, A-150, PC-59, VOC-20, respectively. This suggests that our model train on a predefined set of labels can effectively generalized to other sets of labels during testing. 

\begin{table*}[ht]
    \centering
    \begin{tabular}{|c|c|c|c|c|c|c|c|}
    \hline
       \textbf{Method} & \textbf{Backbone} & \textbf{Training dataset}  & \textbf{A-847} & \textbf{P-459} & \textbf{A-150} & \textbf{PC-59} & \textbf{VOC-20} \\
        \hline
        \hline
        LSeg+ \cite{ghiasi2021open} & R101 & COCO-Panoptic & 2.5 & 5.2 & 13.0 & 36.0 & 59.0\\
        \hline
        \hline
        Simple \cite{xu2021simple} & R101c & COCO-Stuff & - & - & 15.3 & - & 74.5\\
        \hline
        Zegformer \cite{ding2022decoupling} & R50 & COCO-Stuff & - & - & 16.4 & - & 80.7\\
        \hline
        OpenSeg \cite{ghiasi2021open} & R101 & COCO-Panoptic & 4.0 & 6.5 & 15.3 & 36.9 & 60.0\\
        \hline
        OpenSeg \cite{ghiasi2021open} & R101 & COCO-Panoptic + Loc. Narr. & 4.4 & 7.9 & 17.5 & 40.1 & 63.8\\
        \hline
        \hline
        Ours & R50 & COCO-Stuff & 4.0 & 8.5 & 17.6 & 47.6 & 85.6\\
        \hline
        Ours & R50 & COCO-Stuff + COCO Caption & \textbf{4.6} & \textbf{9.6} & \textbf{19.4} & \textbf{48.0} & \textbf{85.9}\\
    \hline
    \end{tabular}
    \caption{Open-vocabulary semantic segmentation results .}
    \label{tab:open_vocab}
\end{table*}

\subsection{Ablation Studies}




\noindent\textbf{Ensemble with CLIP.}
Following previous works \cite{xu2021simple, ding2022decoupling}, we also ensemble predictions from our model with that from the pretrained CLIP model. We use the same ensemble technique in \cite{xu2021simple}. As shown in Table \ref{tab:clip_ensemble}, the ensemble improves the performance in all metrics on the VOC-20 and COCO-Stuff datasets. However, the relative mIoU gains for unseen classes are higher than the gains for seen classes. These results demonstrate that the CLIP ensemble can improve the performance, especially with less training labels.

\begin{table}[ht]
    \centering
    \begin{tabular}{|c|c|c|c|c|}
    \hline
       \textbf{Dataset}  & \textbf{Ensemble} & \textbf{mIoU(S)} & \textbf{mIoU(U)} & \textbf{hIoU}\\
       \hline
       VOC-20 & x & 82.7 & 62.2 & 71.0 \\
       {} & \checkmark & \textbf{88.5} & \textbf{74.8} & \textbf{81.1} \\
       \hline
       COCO  & x & 37.2 & 35.5 & 36.4\\
       Stuff & \checkmark & \textbf{38.6} & \textbf{42.0} & \textbf{40.2} \\
    \hline
    \end{tabular}
    \caption{Results of the ensemble with CLIP on COCO Stuff dataset.}
    \label{tab:clip_ensemble}
\end{table}

\noindent\textbf{Availability of unseen labels.} We compare the results on COCO-Stuff dataset under 3 scenarios: (1) No unseen labels for each image, (2) pseudo unseen labels for each image, and (3) ground truth unseen labels for each image. The numbers are shown in Table \ref{tab:unseen_avail}. The model trained with only seen classes (first row) achieve the best performance on seen classes, but the worst performance on unseen classes. When using pseudo unseen labels for each image, the mIoU for unseen classes gains significantly by 26.9\%. Using ground truth unseen label give the best performance on unseen classes. Note that, all scenarios are image-level labels, not mask annotations.

\begin{table}[ht]
    \centering
    \begin{tabular}{|c|c|c|c|}
    \hline
       \textbf{Variants}  & \textbf{mIoU(S)} & \textbf{mIoU(U)} & \textbf{hIoU}\\
       \hline
       No Unseen & \textbf{40.1} & 25.3 & 31.0 \\
       \hline
       Pseudo Unseen  & 38.6 & 42.0 & 40.2\\
       \hline
       GT Unseen & 39.0 & \textbf{52.2} & \textbf{44.7} \\
    \hline
    \end{tabular}
    \caption{Results under different availability scenarios of unseen labels.}
    \label{tab:unseen_avail}
\end{table}

\noindent\textbf{Benefits of each component.}
We provide an ablation study on each component of the proposed method. In Table \ref{tab:ablation}, CE, KL, R and P are cross entropy loss, KL loss, multi-label ranking loss and text prompt, respectively. All models in this ablation study are trained with ground truth unseen labels for each image. As shown in Table \ref{tab:ablation}, the first row is the baseline with CE loss and `background' embedding. When replacing the `background' embedding with KL loss, the mIoU for unseen classes is increased by 3.7\% because proposals from different unseen classes are not classified to the same `background' class. When using multi-label ranking loss, the mIoU for seen classes is increased by 1.7\%, and the mIoU for unseen classes is significantly improved by 15\%. This shows that class relationship is helpful for both seen and unseen classes. When using text prompt, all the results are further improved.

\begin{table}[ht]
    \centering
    \begin{tabular}{|c|c|c|c|c|c|c|}
    \hline
       \textbf{CE}  & \textbf{KL} & \textbf{R} & \textbf{P} & \textbf{mIoU(S)} & \textbf{mIoU(U)} & \textbf{hIoU}\\
       \hline
       \checkmark & & & & 36.7 & 35.2 & 35.9\\
       \checkmark & \checkmark & & & 31.8 & 38.9 & 35.0\\
       \checkmark & \checkmark & \checkmark & & 38.4 & 50.1 & 43.5\\
       \checkmark & \checkmark & \checkmark & \checkmark & \textbf{39.0} & \textbf{52.2} & \textbf{44.7}\\
    \hline
    \end{tabular}
    \caption{Ablation studies of main components. `CE' stands for cross-entropy loss, `KL' means the KL loss of `background' proposals, `R' is the multi-label ranking loss, and `P' represents prompt templates for text embeddings. Results on COCO Stuff dataset.}
    \label{tab:ablation}
\end{table}

\noindent\textbf{Visualization analysis.} Figure \ref{fig:vis1} shows visualized results of our proposed method and the baseline method using cross-entropy loss with the `background' embedding. Our method successfully classifies unseen `river' and `grass' objects, while the baseline model mis-classifies the `river' as a seen class `water-other' in the top row, and cannot segment the `grass' region in the bottom row.



\section{Conclusion}
In this work, we have identified two problems in the latest mask proposal based ZS3 models: the confusing `background' class and missing the consideration of the label relationships. To address the two problems, we have respectively proposed a KL divergence loss for `background' proposals and a multi-label ranking loss to capture semantic relationship among the proposals. A pseudo label generation pipeline was proposed to further capture the co-occurence relationship between seen and unseen classes.
Experiments on several benchmark datasets show that our method significantly outperforms other methods on GZS3 and OVS.



{\small
\bibliographystyle{unsrtnat}
\bibliography{references}
}







\end{document}